\documentclass[10pt,twocolumn,letterpaper]{article}

\usepackage[pagenumbers]{iccv} 
\usepackage{amsthm}
\usepackage{tcolorbox}

\definecolor{iccvblue}{rgb}{0.21,0.49,0.74}
\usepackage[pagebackref,breaklinks,colorlinks,allcolors=iccvblue]{hyperref}

\theoremstyle{plain}
\newtheorem{theorem}{Theorem}[section]
\newtheorem{proposition}[theorem]{Proposition}

\theoremstyle{definition}

\theoremstyle{remark}

\def\paperID{13301} 
\def\confName{ICCV}
\def\confYear{2025}

\newcommand{\f}{\mathbf{f}}
\newcommand{\x}{\mathbf{x}}

\renewcommand{\b}{\mathbf{b}}
\newcommand{\w}{\mathbf{w}}
\newcommand{\sinc}{\mbox{sinc}}

\newcommand{\I}{\mathbf{I}}

\newcommand{\y}{\mathbf{y}}
\newcommand{\A}{\mathbf{A}}
\newcommand{\U}{\mathbf{U}}
\newcommand{\V}{\mathbf{V}}
\renewcommand{\u}{\mathbf{u}}
\newcommand{\s}{\mathbf{s}}
\newcommand{\diag}{\mbox{diag}}
\newcommand{\ReLU}{\mbox{ReLU}}
\newcommand{\qsection}[1]{\vspace{0.2cm} \noindent \textbf{#1}:}

\renewcommand{\v}{\mathbf{v}}

\title{Gradient Descent as a Shrinkage Operator for Spectral Bias}

\author{Simon Lucey\\
Australian Institute for Machine Learning (AIML)\\
University of Adelaide\\
{\tt\small simon.lucey@adelaide.edu.au}
}

\begin{document}
\maketitle

\begin{abstract}
We generalize the connection between activation function and spline regression/smoothing and characterize how this choice may influence spectral bias within a 1D shallow network. We then demonstrate how gradient descent (GD) can be reinterpreted as a shrinkage operator that masks the singular values of a neural network's Jacobian. Viewed this way, GD implicitly selects the number of frequency components to retain, thereby controlling the spectral bias. An explicit relationship is proposed between the choice of GD hyperparameters (learning rate \& number of iterations) and bandwidth (the number of active components). GD regularization is shown to be effective only with monotonic activation functions. Finally, we highlight the utility of non-monotonic activation functions (sinc, Gaussian) as iteration-efficient surrogates for spectral bias.      
\end{abstract}

\section{Introduction}
Spectral bias -- the tendency of neural networks to prioritize learning low frequency functions~\cite{fridovich2022spectral} -- is often touted as a potential explanation for the extraordinary generalization properties of modern deep neural networks. Let's consider the estimation $f(x)$ of a continuous signal $y(x)$ using a finite number of samples $\mathbb{D}$ such that,  
\begin{equation}
\arg \min_{f} \frac{1}{2} \sum_{\hat{x} \in \mathbb{D}} ||y(\hat{x}) - f(\hat{x})||^{2}  
\label{Eq:error}
\end{equation}
where $f$ is parameterized using a so called \emph{shallow} network, 
\begin{equation}
    f(x) = \sum_{m=0}^{M-1} w_{m} \cdot \eta(x - b_{m}) 
\label{Eq:shallow}
\end{equation}
$\w = [w_{0},\ldots,w_{M-1}]^{T}$ are the network weights, and $\b = [b_{0},\ldots,b_{M-1}]^{T}$ are the bias terms that shift the activation function $\eta(x)$. $M$ is typically referred to as the width of the network. For the purposes of this paper we will assume we are optimizing Eq.~\ref{Eq:error} with gradient descent (GD) as is common practice. When using such networks for signal reconstruction we are often faced with seemingly adhoc design questions around: (i) the type of activation function, (ii) the width of the network, and (iii) the choice of learning rate and number of iterations. Our goal in this paper is to better understand how and when these choices explicitly effect spectral bias.  

\begin{figure}[!htb]
    \centering
    \includegraphics[width=\linewidth]{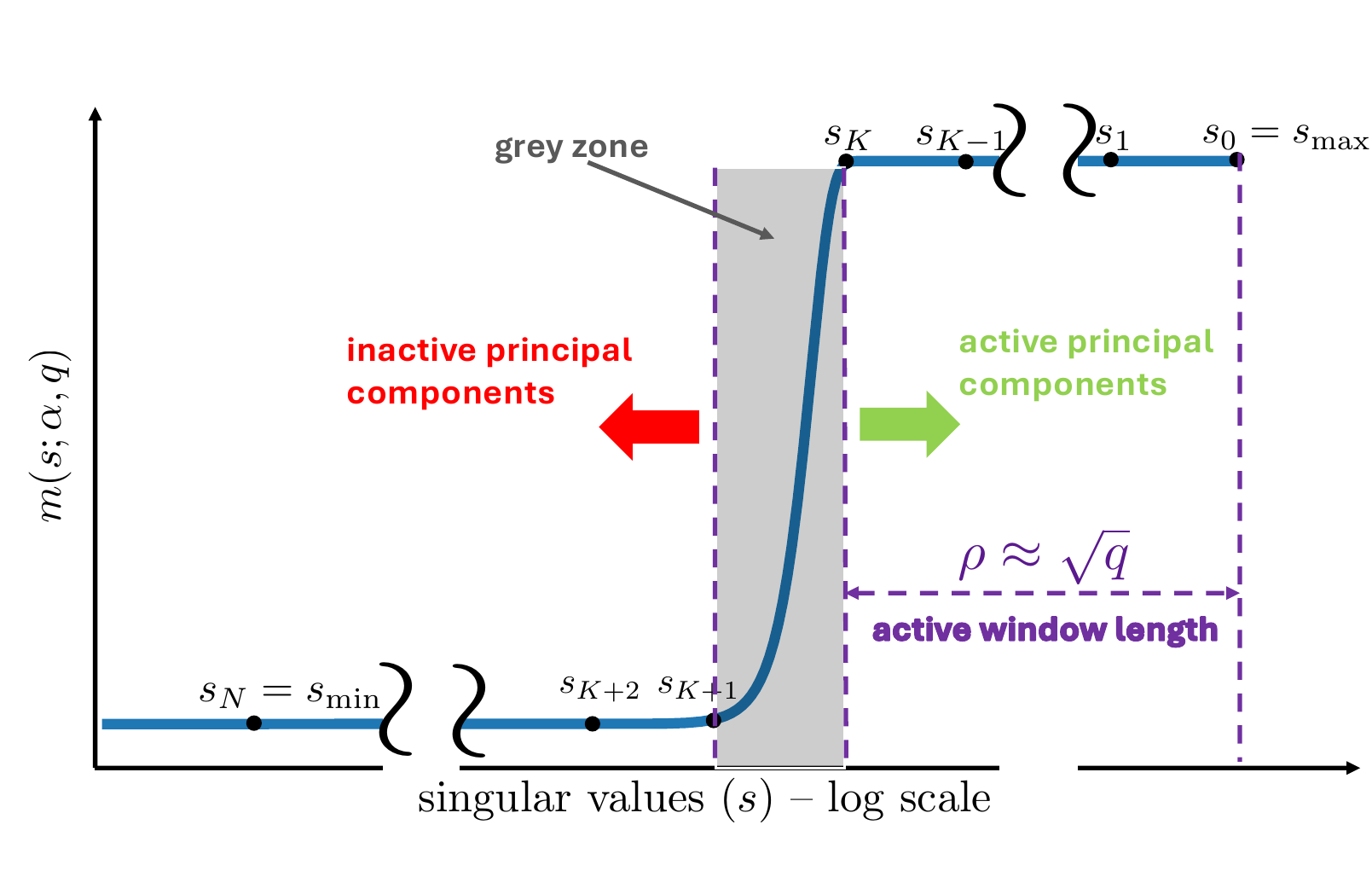}
    \caption{We argue that the implicit regularization offered by gradient descent (GD) can be viewed as a shrinkage operator on the singular values $\s = [s_{0},\ldots,s_{N}]$ of a network's Jacobian $\nabla f(\x)$. The operator applies a masking function $m(s;\alpha,q)$ on the singular values, where $\alpha$ is learning rate and $q$ is the number of iterations which dictate the number $K$ of active principal components. The distribution of $\s$ is activation specific, and therefore how effectively it can be regularized through GD.}  
    \label{fig:intro}
\end{figure}


\vspace{0.2cm}
\noindent \textbf{Contributions:}
In Sec.~\ref{Sec:impulse} we generalize the connection between activation function choice and the minimization of gradient energy in spline regression/smoothing. We then show in Sec.~\ref{Sec:GD} how GD can be viewed as a shrinkage operator on the singular values of a network's Jacobian $\nabla f(\x)$ (see Fig.~\ref{fig:intro}). Further, we show that activation functions (Heaviside, tanh) have singular values that relate directly to a frequency basis -- specifically the discrete sine transform (DST). For modern activations (e.g. ReLU, GELU, SiLU) the approximation to the frequency basis becomes less strict but is still apparent. An explicit relationship between the number of active principal components and GD hyper-parameters is then established. A heuristic is proposed between (learning rate, number of iterations) and spectral-bandwidth for activations. We are the first, to our knowledge, to make explicit such a relationship. Finally, we demonstrate that the manipulation of (learning rate, number of iterations) -- as a strategy for implicit regularization -- is only effective for monotonic activations. Non-monotonic activations (sinc, Gaussian, etc.) instead rely on scaling as their method for implicit regularization. We highlight sinc in particular as an iteration efficient surrogate for enforcing spectral bias.    

\section{Background}
For the case of least-squares it is understood that GD will converge to the minimum Euclidean norm solution~\cite{soudry2018implicit}. The interplay between activation function and GD hyper-parameter choice, however, in this regularization process is not as well understood for neural networks, and is the central focus of this paper. 

\qsection{Activation bias}
ReLU networks tend to learn low frequency components first when being trained using GD. Sinusoid activations were proposed~\cite{tancik2020fourier,sitzmann2020implicit} to allow these networks to model higher bandwidth signals. NTK analysis~\cite{jacot2018neural} was used by Tancik et al.~\cite{tancik2020fourier} to explain this behaviour and motivate the use of a sinusoid activated first layer -- referred to as Fourier positional encoding -- within a deep ReLU network. Sitzman et al.~\cite{sitzmann2020implicit} then demonstrated that a fully sinusoid activated deep network -- suitably initialized -- can achieve even better performance. Ramasinghe et al.~\cite{ramasinghe2022beyond} argued that exponential family activations like Gaussian can achieve similar performance without requiring specialized initialization. Recently, it has been noted empirically, that sinc~\cite{saratchandransampling} activations can achieve even better results. A focus of this paper is to make explicit this spectral bias for the case of a 1D shallow network. 

\qsection{GD vs. SGD}
In spite of previous works~\cite{charles2018stability,hardt2016train} stating that stochastic sampling is essential -- many empirical studies now exist that show full-batch GD optimized networks generalize well. Hoffer et al.\cite{hoffer2017train} demonstrated with a sufficient number of iterations, full-batch GD generalizes similarly to SGD. Likewise, Geiping et al.\cite{geipingstochastic} empirically found that generalization behavior persists even without stochastic sampling. For smooth losses -- such as the $\ell_{2}$ loss been considered in Eq.~\ref{Eq:error} -- it has been shown~\cite{nikolakakisbeyond} that GD exhbits tighter generalization bounds than SGD and therefore will be of central focus in this paper. 

\qsection{ReLU dilemma}
A classic activation used throughout modern neural network literature~\cite{krizhevsky2012imagenet} is $\ReLU(x) = \max(x,0)$. What has surprised much of the ML community in recent years, however, is how poorly this popular activation performs for low dimensional signal reconstruction problems like the one described in Eq.~\ref{Eq:error}. What makes this even more confusing is seminal work by Zhang et al.~\cite{zhang2017understanding} who proved that a 1D shallow ReLU network like the one described in Eq.~\ref{Eq:shallow} could perfectly reconstruct training samples $\mathbb{D}$ provided the width and distribution of $\b$ is sufficient. A drawback to this analysis, however, is that it ignores the optimization process required to find the solution -- namely GD. 

\section{Smoothness \& Activations}
\label{Sec:impulse}
Smoothness stems directly from the well established framework of empirical risk minimization (ERM)~\cite{vapnik1991principles}, and has been used since the earliest days of ML to quantify generalization (in regression). In summary, given a set of hypotheses (models) that minimizes the empirical risk (with training data), the ERM framework prefers a solution that minimizes the true risk (with respect to the actual data distribution) with a higher probability. When extra prior knowledge is unavailable on the true data distribution, ERM suggests that the best solution would be the one that minimizes the least complex solution that minimizes the empirical risk (under the realizability assumption). This can be primarily achieved through: (i) regularizing the parameters of the model or (ii) regularizing the function output itself. Popular regularizations such as ridge regression fall into the first category, whereas spline regression with regularized derivatives fall into the second category. 

\qsection{Connecting (i) and (ii)}
 Spline regression/smoothing~\cite{reinsch1967smoothing} makes no parametric assumption on $f$, instead it adds an explicit regularization of the $r$-th order gradient over the input domain, 
\begin{equation}
\arg \min_{f} \frac{1}{2} \sum_{\hat{x} \in \mathbb{D}} ||y(\hat{x}) - f(\hat{x})||^{2} + \frac{\lambda}{2} \int_{x} ||\frac{d^{r}}{d x^{r}} f(x)||^{2} dx
\label{Eq:spline_error}
\end{equation}
where $r$ is the order of the gradient, and $\lambda$ is a non-negative regularization factor that controls the amount of smoothness. Typically the second order gradient (i.e. $r = 2$) is minimized, but other variations can be entertained. Intuitively, setting $\lambda$ to a larger value will make $f$ smoother and more flat, whereas keeping $\lambda$ smaller preserves detail but risks the introduction of discontinuities.  

Yuille and Grzywacz~\cite{yuille1989mathematical} highlighted the theoretical connection between regularization strategies (i) and (ii) in their seminal work on motion coherence theory within computer vision. A connection with spline smoothing using second order gradients and ReLU activations in a shallow network was recently established~\cite{williams2019gradient} and \cite{heiss2019implicit}. We establish the following more general connection, 
\begin{proposition}
The spline regression/smoothing objective in Eq.~\ref{Eq:spline_error} can be equivalently expressed as, 
\begin{eqnarray}
\arg \min_{w} & & \frac{1}{2} \sum_{\hat{x} \in \mathbb{D}} ||y(\hat{x}) - f(\hat{x};w)||^{2} + \frac{\lambda}{2} \int_{x} ||w(x)||^{2} dx \nonumber \\
\mbox{s.t.} & & f(\hat{x};w) = \int_{x} w(x) \eta(\hat{x} - x) dx  
\label{Eq:activation_cont}
\end{eqnarray}
where $\eta(x) = x^{r-1} \cdot [x > 0]$ for $r \geq 1$.
\label{Prop:Act}
\end{proposition}
A full derivation and empirical analysis of this proposition can be found in Appendix~\ref{App:hside}. In Eq.~\ref{Eq:activation_cont} we allow an abuse of notation and allow $w$ to be a continuous function. Inspecting the objective we now have the familiar ridge regression penalty on $w$ as advocated for in (i). Both Williams et al. and Heiss et al. demonstrated (independently) that for second order gradients ($r = 2$) the ReLU activation $\eta(x) = x \cdot [x > 0] = \max(x,0)$ enforces the same smoothing. To our knowledge no-one has previously established the more general connection for the $r-$th order gradient penalty using our proposed reformulation in Eq.~\ref{Eq:activation_cont}. Using this result it is trivial to show that when minimizing first order gradients ($r=1$) the equivalent activation becomes a Heaviside function~$\eta(x) = [x > 0]$, further when ($r=2$) the activation becomes ReLU2 $\eta(x) = x^{2} \cdot [x > 0]$. 

\qsection{Continuous to discrete}
It is now easy to see that the shallow network described in Eq.~\ref{Eq:shallow} can be seen as a discrete approximation to the spline regularized objective described in Eq.~\ref{Eq:activation_cont}. The choice of activation implicitly controls the type of gradient smoothing being enforced. Specifically, the approximation strictly holds if we replace the continuous integral with a numerical one $\sum_{j=0}^{M-1} w_{j} \eta(\hat{x} - b_{j})$ which becomes more accurate as $M \rightarrow \infty$. Ideally, we would choose $\b = [b_{j}]_{j}$ to have $M$ equal spaces between zero and $T$, but other distributions can be entertained so long the minimum distance between biases is below a desired threshold. The continuous function $w$ is now replaced by the discrete vector $\w = [w_{j}]_{j}$. Heiss et al. reported a similar result for $r=2$ stating that the spline approximation strictly holds as the width of the shallow network becomes infinite.

\section{Gradient Descent as a Regularizer}
\label{Sec:GD}
We can combine Eqs.~\ref{Eq:error} and~\ref{Eq:shallow} into vector form so we are minimizing,
\begin{equation}
    \w^{*} = \arg \min_{\w}  \mathcal{E}(\w) + \Omega(\w)
    \label{Eq:ls}
\end{equation}
where
\begin{equation}
    \mathcal{E}(\w) = \frac{1}{2}||\y - \A \w||_{2}^{2}
    \label{Eq:Ew}
\end{equation}
such that $\y = [y(\hat{x})]_{\hat{x} \in \mathbb{D}}$, $\hat{\x} = [\hat{x}]_{\hat{x} \in \mathbb{D}}$, $N = |\mathbb{D}|$ and $\A = [\eta(\hat{x}_{i} - b_{j})]_{ij}$ forms a $N \times M$ matrix. Lets also assume that $M > N$ so that the solution $\mathcal{E}(\w)$ is not unique unless an additional regularization penalty $\Omega(\w)$ is employed. As discussed a classical choice for the regularization term in Eq.~\ref{Eq:ls} is ridge regularization $\Omega(\w) = \frac{\lambda}{2}||\w||_{2}^{2}$. It should also be noted that one can think of the matrix $\A^{T}$ formally as the shallow network's Jacobian $\nabla f(\hat{\x})$ where we are only optimizing for the weights $\w$. For most of this paper we will assume that $\b$ is known, although strategies for choosing $\b$ will be discussed.

\qsection{Shrinkage operator}
If we were to solve the objective (Eq.~\ref{Eq:ls}) in closed form using an SVD this is equivalent to applying a shrinkage operator to the singular values of $\A$ such that,
\begin{equation}
     \w^{*} = \V [\diag(\hat{\s}^{-1})] \U^{T} \y
     \label{Eq:w}
\end{equation}
where 
\begin{equation}
    \hat{s}_{\mbox{\tiny{rdg}}}^{-1} = \frac{s}{s^{2} + \lambda}
    \label{Eq:ridge}
\end{equation}
and $\A = \U [\mbox{diag}(\s)] \V^{T}$ is the SVD decomposition, $\U$ and $\V$ are orthobases and $\s$ is a vector of non-negative singular values. Eq.~\ref{Eq:ridge} is applied to each element of $\s$ to generate $\hat{\s}_{\mbox{\tiny{rdg}}}^{-1}$ from which $\w^{*}$ is obtained by applying Eq.~\ref{Eq:w}. 

\qsection{PCA regularization}
A drawback to ridge regularization, however, is that it attenuates all the singular components of $\A$, but has no mechanism to actively select (with no attenuation) the first $K$ principal components. An alternative known as PCA regularization/regression~\cite{jolliffe1982note} is sometimes preferred in this regard, 
\begin{equation}
    \hat{s}_{\mbox{\tiny{pca}}}^{-1} = m_{\mbox{\tiny{pca}}}(s,\kappa) \cdot s^{-1}
    \label{Eq:spca}
\end{equation}
and
\begin{equation}
    m_{\mbox{\tiny{pca}}}(s;\kappa) = 
    \begin{cases}
      1, & \text{if $s \geq \kappa$} \\
      0, & \text{otherwise.}
    \end{cases}      
    \label{Eq:mask_pca}
\end{equation}
The threshold $\kappa$ can be tuned as a hyperparameter to select only the first $K$ principal components of $\A$. This is also referred to in literature~\cite{gavish2017optimal} as hard-threshold shrinking.   

\qsection{GD regularization}
Well regularized solutions to the least-squares problem described in Eq.~\ref{Eq:ls} can be achieved using GD without the need for an explicit regularizer. This instead involves solving the problem in an iterative fashion such that,
\begin{equation}
\w \rightarrow \w
 - \alpha \cdot \frac{\partial \mathcal{E}(\mathbf{w})}{\partial \mathbf{w}^{T}}
\label{Eq:GD1}
\end{equation}
where $\alpha$ is the learning rate and
\begin{equation}
\frac{\partial \mathcal{E}(\mathbf{w})}{\partial \mathbf{w}^{T}} = -\A^{T}[\y - \A \mathbf{w}].
\label{Eq:GD2}
\end{equation}

\begin{figure}[!htb]
    \centering
    \includegraphics[width=\linewidth]{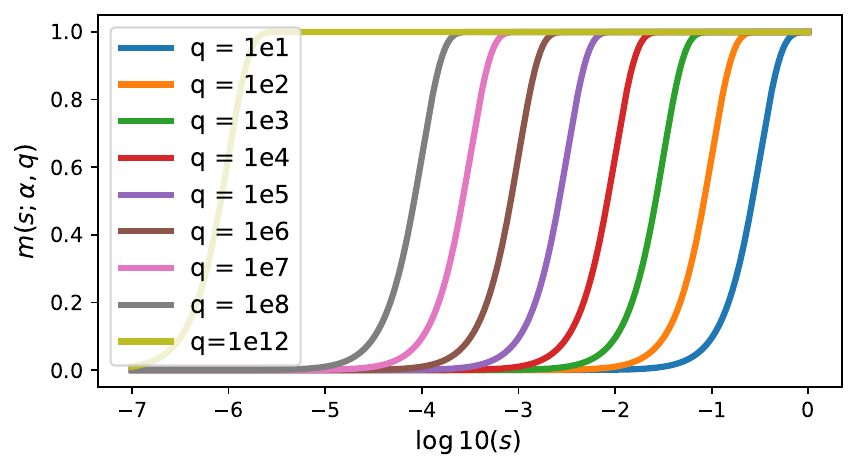}
    \caption{Visual depiction of the masking function $m(s; \alpha, q)$ used for shrinking singular values within GD regularization. We set $\alpha = 1$ so as to ensure the upper singular value of the mask is normalized at unity. We then vary the number of iterations $q$ to demonstrate how this effects the width $\rho$ of the active window. }
    \label{Fig:mask}
\end{figure}

Gradient flows~\cite{jacot2018neural} is a well studied approach for approximating GD as a continuous \emph{ordinary differential equation} (ODE)
\begin{equation}
    \frac{\partial \w}{\partial t} = \lim_{\alpha \rightarrow 0} \frac{\w_{q+1} - \w_{q}}{\alpha} = -\frac{ \partial \mathcal{E}( \w )}{\partial \w^{T}}
\end{equation}
where $t$ is a continuous variable modeling how the ODE evolves over time. It is also often used to model -- in closed form -- the approximate regularization effects of GD. The approach has been used to great effect within NTK theory~\cite{jacot2018neural} to approximate the training of neural networks as a linear kernel regression. 
\begin{proposition}
The implicit regularization of GD when minimizing the least-squares problem in Eq.~\ref{Eq:Ew} can be reinterpreted as an explicit shrinkage operator on the singular values of $\A$ -- with  iterations $q$, and learning rate $\alpha$ -- such that,
\begin{equation}
    \hat{s}_{\mbox{\tiny{gd}}}^{-1} = m_{\mbox{\tiny{gd}}}(s; \alpha, q) \cdot s^{-1}
    \label{Eq:sgd}
\end{equation}
where, 
\begin{equation}
m_{\mbox{\tiny{gd}}}(s; \alpha, q) = [1 - \exp(-\alpha \cdot q \cdot s^{2})]  \;\;. 
\label{Eq:mask_gd}
\end{equation}
\label{Prop:GD}
\end{proposition}
A full derivation can be found in Appendix~\ref{App:GD}. Here one can see that iterations $q$, and learning rate $\alpha$ replace $\lambda$ (in ridge shrinkage) as the regularizing hyper-parameters. A visualization of the masking function in Eq.~\ref{Eq:mask_gd} can be found in Fig.~\ref{Fig:mask}. One can also see the masking function operates in a similar manner to the one employed with PCA regularization in Eq.~\ref{Eq:mask_pca} in that it can be used to select which singular values are active. PCA regularization uses a single parameter $\kappa$ to select active singular values. GD uses two parameters: (i) learning rate $\alpha$ which dictates how the masking function is shifted, and (ii) number of iterations $q$ which controls the length of the active window $\rho$ (see Fig:~\ref{fig:intro}). We can relate $\kappa$ with $(\alpha,q)$ explicitly, 
\begin{proposition}
A relationship between the singular value threshold $\kappa$, learning rate $\alpha$, and the number of iterations $q$ can be made, 
\begin{equation}
    \kappa = \sqrt{\frac{-\log(\epsilon)}{\alpha \cdot q}}
\end{equation}
where the mask threshold $\epsilon$ determines the singular value $\kappa$ where $m_{\tiny{\mbox{gd}}}(\kappa; \alpha, q) = 1 - \epsilon$. The above relation simplifies down to, 
\begin{equation}
q = s^{2}_{\max} \cdot s_{K}^{-2} 
\label{Eq:q}
\end{equation}
if we choose learning rate $\alpha = s^{-2}_{\max}$, mask threshold $\epsilon = \exp(-1)$ and singular value threshold $\kappa = s_{K}$.
\label{Prop:q}
\end{proposition}
We should note that a learning rate $\alpha < s^{-2}_{\max}$ can be entertained, but will then require an increase in $q$ to maintain the same singular value threshold $\kappa$. From an iteration efficiency perspective we therefore use $\alpha = s^{-2}_{\max}$ in Eq.~\ref{Eq:q}. For PCA we must select $s_{K} > \kappa \geq s_{K+1}$, but due to the slow mask transition of GD it is sufficient to set $\kappa = s_{K}$ (see next section). Arbitrary values of the mask threshold can also be entertained $1 > \epsilon > 0$, $\epsilon = \exp(-1)$ was chosen to simplify the exposition of the relationship of factors. A full derivation of Proposition~\ref{Prop:q} can be found in Appendix~\ref{App:rho}. We shall use the relation in Eq.~\ref{Eq:q} as a mechanism for GD to choose the number of principal components $K$ to implicitly preserve through $(\alpha,q)$. The width of the active window for the mask function (see Fig~\ref{fig:intro}) can also be determined through this relation $\rho = s_{\max} \cdot s_{K}^{-1} = \sqrt{q}$.

\qsection{PCA vs. GD}
There are, however, two clear differences between PCA and GD regularization. First, unlike PCA regularization the active window -- that is the region of the masking function that is at unity -- is of a finite width. Through visual inspection of Fig.~\ref{Fig:mask} one can see that the square of the window width -- which we refer to as $\rho$  -- is approximately equal to the number of iterations $q$. Second, the ability for GD to switch singular values on and off through its masking function is quite poor. We found a drop off of approximately $15$dB to complete 99\% of the transition from one to zero. We referred to this portion of the masking function -- lying between zero and one -- as the ``grey zone'' in Fig.~\ref{fig:intro} . This is in stark contrast to PCA regularization which enjoys an instantaneous transition. The inability of GD to instantaneously switch singular values on and off is critical for understanding the limits of implicit regularization and choice of activation. 

\section{Memorization \& Gradient Descent}
\label{Sec:MemGD}
The choice of activation also effects the memorization properties of a shallow network (as described in Eq.~\ref{Eq:shallow}) when using GD. A simple solution is
\begin{equation}
    f(x) = \sum_{m = 0}^{M - 1} y(b_{m}) \cdot \delta(x - b_{m})
    \label{Eq:impulse}
\end{equation}
where $\delta(x)$ is an impulse activation function, $\b = [b_{0},\ldots,b_{M-1}]$ are samples drawn from $\mathbb{D}$ such that $M = |\mathbb{D}|$. 
An advantage of this solution is that it is guaranteed of achieving perfect signal reconstruction irrespective of the complexity of the signal $y$ or the number of samples in $\mathbb{D}$. The memorization is obvious since $\A = [\delta(\hat{x}_{i} - \hat{x}_{j})]_{ij}$ is a $N \times N$ identity matrix therefore $\w^{*} = \y$. 

This property does not hold just for impulse activations, but can extend to a wide array of activations so long as $\mbox{rank}(\A \A^{T}) = N$ where $\A = [\eta(\hat{x}_{i} - b_{j})]_{ij}$ and $\b$ is an $M \geq N$ dimensional bias vector. Of particular note in this regard is Zhang et. al~\cite{zhang2017understanding} which demonstrated for ReLU that a full rank matrix $\A$ can be realized provided sufficient width and distribution of $\b$. Similar formal proofs can be applied to other activations to demonstrate their ability to memorize~\cite{cybenko1989approximation,gomes09}. In practice, however, it is not the rank of $\A \A^{T}$ that matters when using GD it is instead the distribution of the singular values. In particular for GD regularization the ratio of the largest singular value $s_{\max}$ to the threshold singular value threshold $\kappa$ -- where $s_{K} > \kappa \geq s_{K+1}$ dictates the $K$ principal components we want to preserve -- is critical. 

\qsection{Non-monotonic activations}
Monotonic activations such as ReLU and its many variants dominate current neural network literature. Non-monotonic activations, however, like Gaussian, sinc and sine have become increasingly popular in implicit neural functions~\cite{sitzmann2020implicit,ramasinghe2022beyond,saratchandransampling}. The sinc activation is of specific interest in this paper as the shallow network described in Eq.~\ref{Eq:shallow} strikes a strong resemblance to Shannon \& Whitaker's classic sampling theorem equation~\cite{shannon1949communication,whittaker1928fourier}
\begin{equation}
    f(x) = \sum_{k \in \mathbb{Z}} f(K^{-1}k) \cdot \sinc(K \cdot [x - K^{-1}k])
    \label{Eq:shannon}
\end{equation}
where $B = K/2$ is considered the \emph{bandwidth} of the signal $f(x)$, and $\sinc(x) = \sin(\pi x) \cdot (\pi x)^{-1}$. Referencing Eq.~\ref{Eq:shallow} $w_{k} = f(K^{-1}k)$ are the weights, and $\eta(x) = \sinc(x)$ is the shifted activation function. This theorem is fundamental to information theory and signal processing as it characterizes a category of continuous signals $f(x)$ that is completely described by $K^{-1}$ spaced discrete samples $f(K^{-1}k)$. The $\sinc$ activation function is optimal in this sense (and forms an orthobasis), but this does not formally hold over a finite spatial extent as it decays slowly spatially~\cite{unser2000sampling}. An often used approximation is the Gaussian activation function  due to its superior spatial decay and computational simplicity (although it does not form an orthobasis). In kernel regression literature the scaling factor~$\sigma$ within the unnormalized Gaussian activation/kernel $\eta(\sigma \cdot x) = \exp(- \frac{\sigma}{\pi} \cdot x^{2})$ is often also referred to as bandwidth, but the connection $\sigma = K$ to formal Shannon \& Whitaker bandwidth is often overlooked\footnote{Yuille and Grzywacz~\cite{yuille1989mathematical} derived a formal connection between the spline regularization term (see Eq.~\ref{Eq:spline_error}), and the bandwidth scaling factor $\sigma$ used in Gaussian activations.}. The re-intepretation of sine activations through the lens of shifted Gaussian activations is now well understood~\cite{zheng2022trading,zheng2021rethinking}.

\qsection{Quasi-monotonic activations}
Modern activations like GELU, SiLU, etc. are not strictly monotonic (i.e. $\eta(x_{2}) \geq \eta(x_{1})$ when $x_{2} > x_{1}$), however, we entertain the label ``monotonic'' as these  activations $\eta(\sigma \cdot x)$ tend towards ReLU (which is monotonic) as $\sigma \rightarrow \infty$. In contrast, non-monotonic activations like sinc and Gaussian tend instead towards an impulse (which is non-monotonic).   

\section{Experiments}
\begin{figure*}[!htb]
    \centering
    \begin{minipage}{.25\textwidth}
        \centering
        \includegraphics[width=\linewidth]{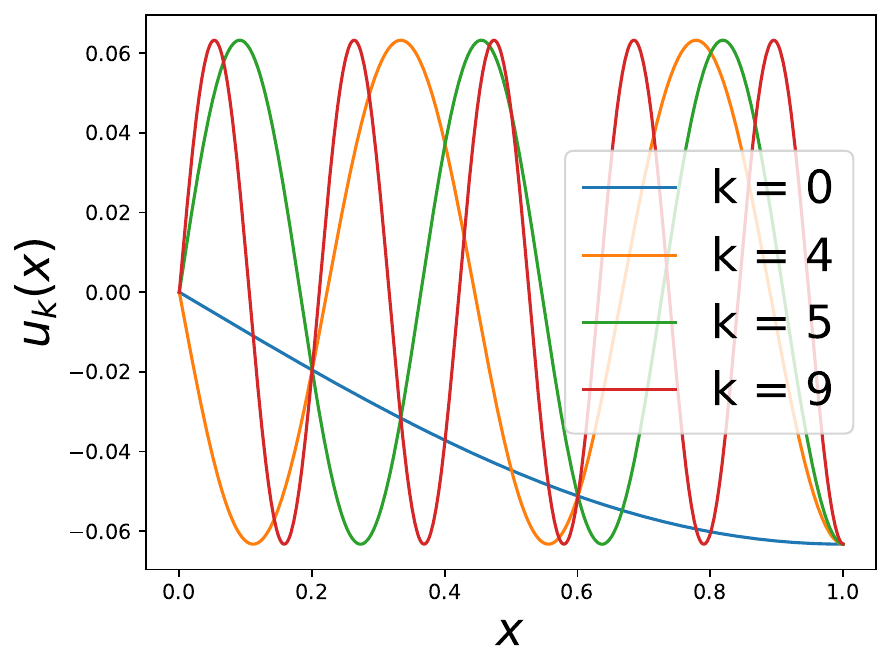}
        (a): Heaviside
    \end{minipage}%
    \begin{minipage}{.25\textwidth}
        \centering
        \includegraphics[width=\linewidth]{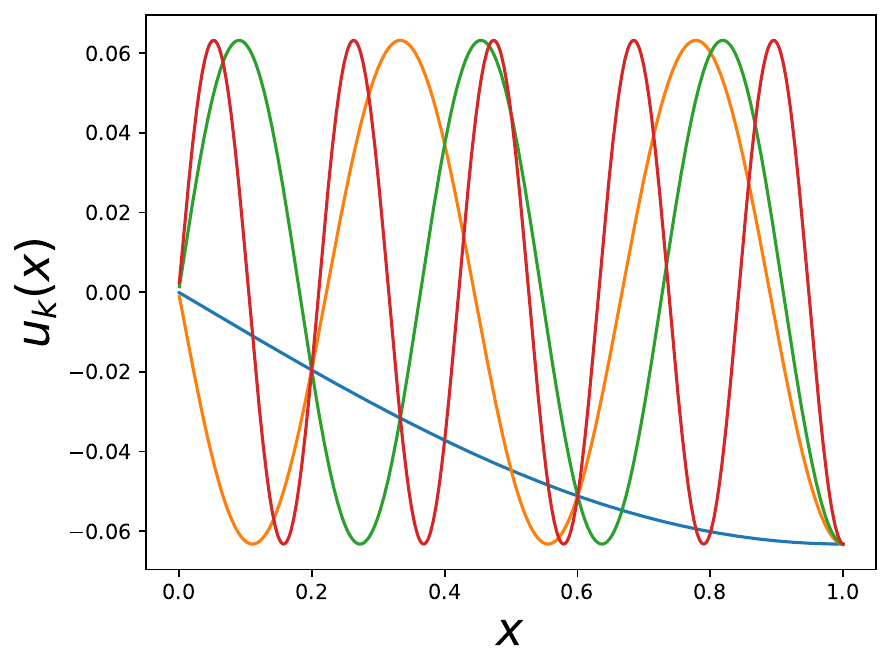}
        (b): tanh
    \end{minipage}%
    \begin{minipage}{.25\textwidth}
        \centering
        \includegraphics[width=\linewidth]{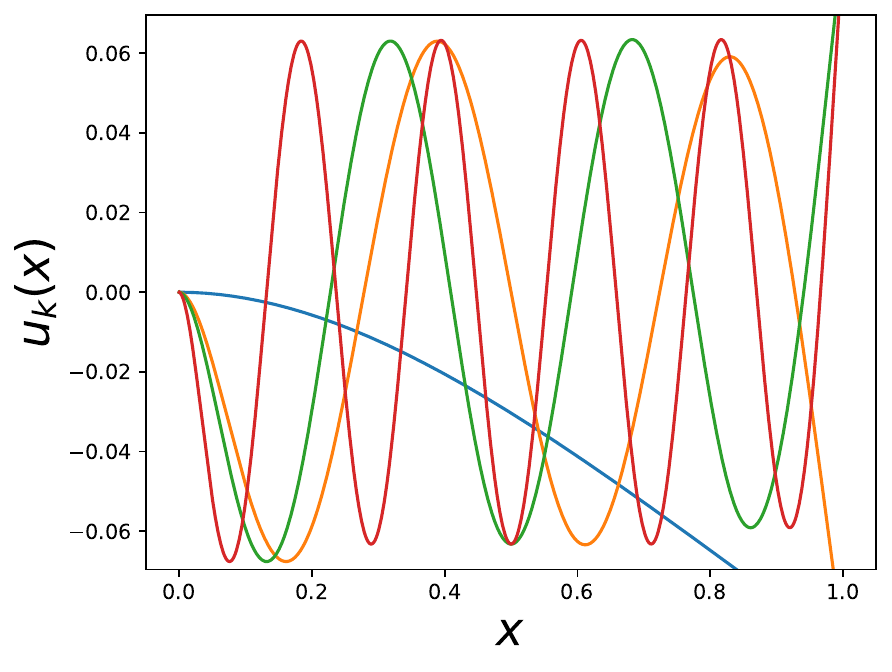}
        (c): ReLU
    \end{minipage}%
    \begin{minipage}{.25\textwidth}
        \centering
        \includegraphics[width=\linewidth]{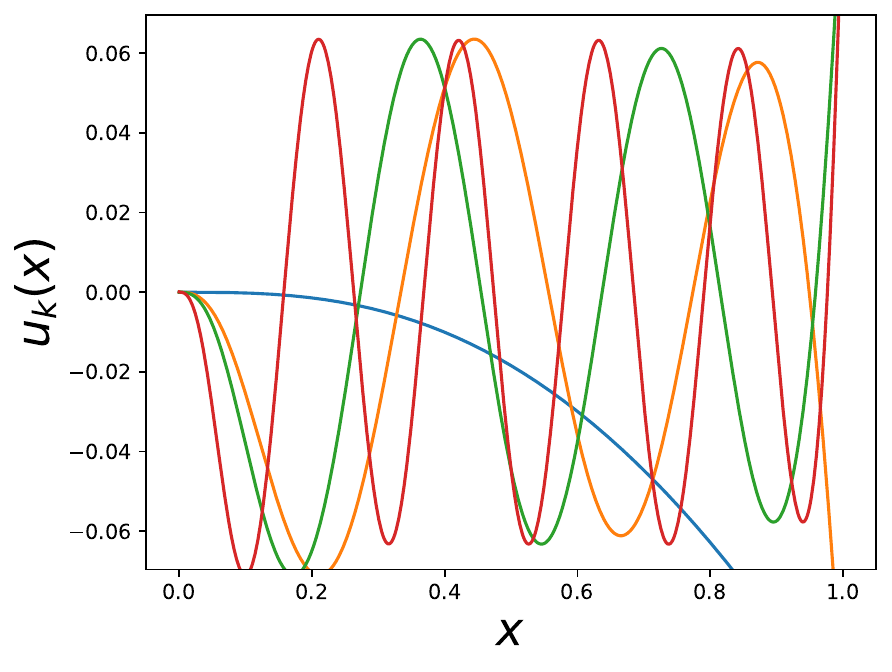}
        (d): ReLU2
    \end{minipage}%
    \caption{Visualization of principal components $u_{k}(x)$ of $\A$ for activation functions (a-d) where $k = [0,4,5,9]$ for $N = M = 5000$. Both (a) Heaviside and (b) tanh exhibit a strict frequency basis, whereas activations like (c) ReLU and (d) ReLU2 have a smooth but less strict frequency basis.}
    \label{Fig:basis_500}
\end{figure*}

\vspace{-3mm}
\qsection{Spectrum of $\A$}
In Fig.~\ref{Fig:basis_500} we can see the SVD distribution of the normalized singular values of $\A = \eta(\sigma \cdot [\x - \b^{T}])$ for: (a) monotonic, and (b) non-monotonic activations. We estimate the distribution numerically by setting $\x$ and $\b$ to have $N$ and $M$ equally spaced values between zero and one (i.e. assuming for $x \in [0,T]$ and that $T=1$). For this particular visualization we have set $M$ and $N$ to $5000$.  

We now want to remind the reader of our insight (see Sec.~\ref{Sec:GD}) that GD implicitly acts as a shrinkage operator on the singular values of the network's Jacobian. In many respects GD behaves like PCA regularization~\cite{jolliffe1982note}, where the iterations $q$ and learning rate $\alpha$ are tuned to preserve principal components. The approximation, however, is not perfect as the masking of the singular values with GD regularization is not instantaneous. The more rapid the drop off between singular values, the better the selectivity of the principal components. 

Inspecting (a) one can see all the monotonic activations have a rapid initial drop-off of singular values -- especially for the early principal components. The rate of singular value drop-off, however, is function of $r$ where $\eta(x) = x^{r-1} \cdot [x > 1]$ (see Proposition~\ref{Prop:Act}). $r=1$ (Heaviside) has the slowest drop-off, then $r=2$ (ReLU), followed by $r=3$ (ReLU2). One can see that tanh tends towards $r=1$, and (GELU, SiLU) tends towards $r=2$. In (b) one can see that the non-monotonic activations have a strikingly different spectral distribution. (sinc, Gaussian) have a tepid then rapid drop-off (at $k = \sigma$). This effect is most extreme for sinc whose singular values stay at unity until $k = \sigma$. We hypothesize that for non-monotonic activations the number of active principal components is predominantly controlled by the manipulation of $\sigma$. In contrast, for monotonic the number of active components is controlled by the learning rate $\alpha$ and iterations $q$. For our visualizations we chose $\sigma = (15,30)$. 

Using Eq.~\ref{Eq:q} we can also relate the number of iterations required to preserve $K$ principal components for each activation (see Appendix~\ref{App:iter}). A real strength of the $r=1$ activations (Heaviside, tanh) is that they need orders of magnitude less iterations to regularize their solution through GD. Conversely, the number of iterations required to regularize an $r=3$ activation such as ReLU2 makes their practical use with GD untenable. 

\begin{figure*}[!htb]
    \centering
    \begin{minipage}{.5\textwidth}
        \centering
        \includegraphics[width=\linewidth]{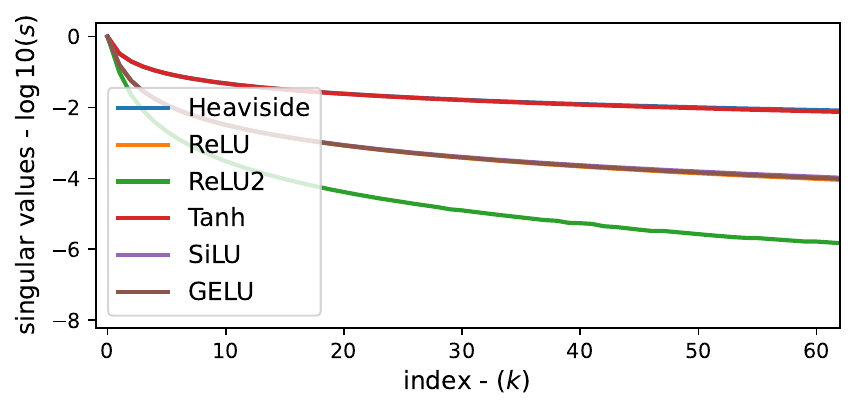}
        (a)
    \end{minipage}%
    \begin{minipage}{.5\textwidth}
        \centering
        \includegraphics[width=\linewidth]{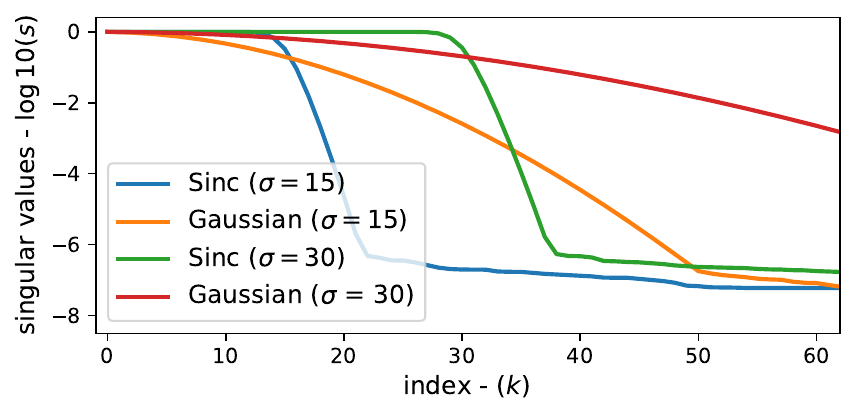}
        (b)
    \end{minipage}%
    \caption{Spectrum of the normalized singular values $\A = \eta(\sigma \cdot [\x - \b^{T}])$ for different activations. In (a) one can see that monotonic activations Heaviside -- $(r=1)$, ReLU -- $(r=2)$, and ReLU2 -- $(r=3)$ -- derived in Proposition~\ref{Prop:Act} -- have different drop off rates related to $r$. tanh has a similar spectrum to Heaviside, whereas GELU and SiLU behave more like ReLU. In contrast to (b) all the monotonic activations enjoy an initial steep drop off, but then plateu. We hypothesize this property to be essential for GD to implicitly regularize for smooth solutions. Conversely, in (b) sinc \& Gaussian plateu early, but then see signficant drop off at $k = \sigma$. This is most noticable for sinc as the normalized singular values remain at unity until $k = \sigma$ followed by a precipitous drop off. Since these non-monotonic activations lack the initial steep drop off they instead rely on the scaling $\sigma$ to promote smooth solutions. We hypothesize that the manipulation of GD hyper-parameters (learning rate, number of iterations) will have little to no regularization effect. It should be noted that $\sigma$ has no effect on the spectral distribution of Heaviside, ReLU and ReLU2 as they are all strictly scale equivariant. We arbitrarily use $\sigma$ values of $15$ \& $30$ in our visualization.}
    \label{Fig:spectrum}
\end{figure*}

\qsection{Principal components of $\A$}
We can decompose $\A = \eta(\sigma \cdot [\x - \b^{T}]) = \U \mbox{diag}(\s) \V^{T}$ using SVD where $\U = [\u_{0}, \ldots, \u_{K}]$ form the principal components of $\A \A^{T}$. In our visualizations we shall approximate the discrete principal component vectors as continuous functions $u_{k}(x) = \u_{k}[x \cdot M \cdot T^{-1}]$. Inspecting Fig.~\ref{Fig:basis_500} we show the principal components for: (a) Heaviside, (b) tanh, (c) ReLU, and (d) ReLU2 activations where $M = N = 5000$. In (a) you can see that the principal components of the Heaviside activation adheres to a variant of the discrete sine transform (DST) frequency basis such that $u_{k}(x) = \vartheta \cdot \sin(\pi \cdot [k + 0.5] \cdot x)$ where $\vartheta$ is an arbitrary scale factor. In (b) tanh exhibits a similar property, although is partially sensitive to choice of scaling. We found $\sigma \approx M$ to be a good empirical choice. The similar performance of (a) and (b) is unsurprising as $\tanh(\sigma \cdot x)$ tends towards a Heaviside function as $\sigma \rightarrow \infty$. In (c) ReLU and (d) ReLU2 both activations form a smooth basis, but becomes more inexact to a frequency basis as $r$ increases where $\eta(x) = x^{r-1} \cdot [x > 0]$ (see Proposition~\ref{Prop:Act}). 

\begin{figure*}[!htb]
    \centering
    \begin{minipage}{.25\textwidth}
        \centering
        \includegraphics[width=\linewidth]{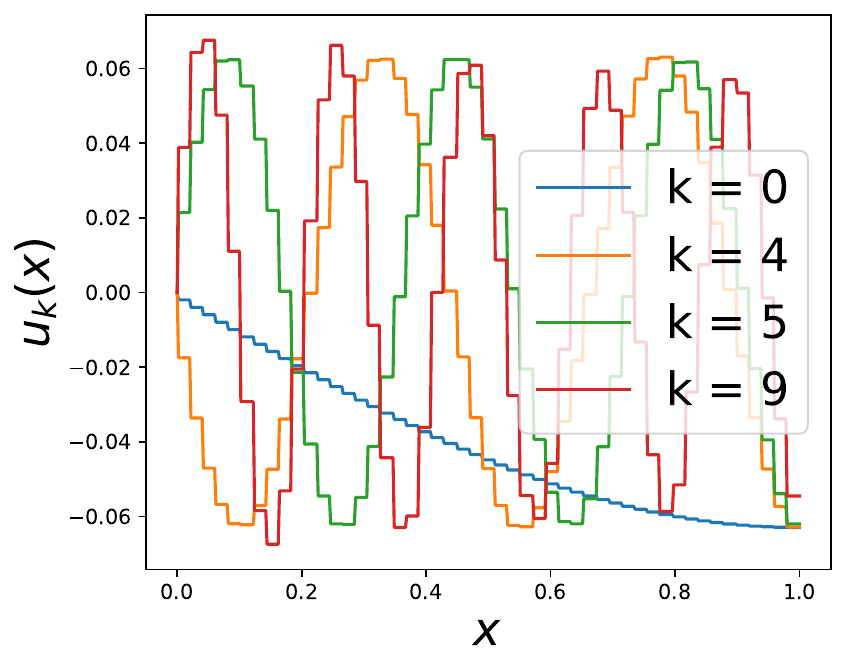}
        (a): Heaviside
    \end{minipage}%
    \begin{minipage}{.25\textwidth}
        \centering
        \includegraphics[width=\linewidth]{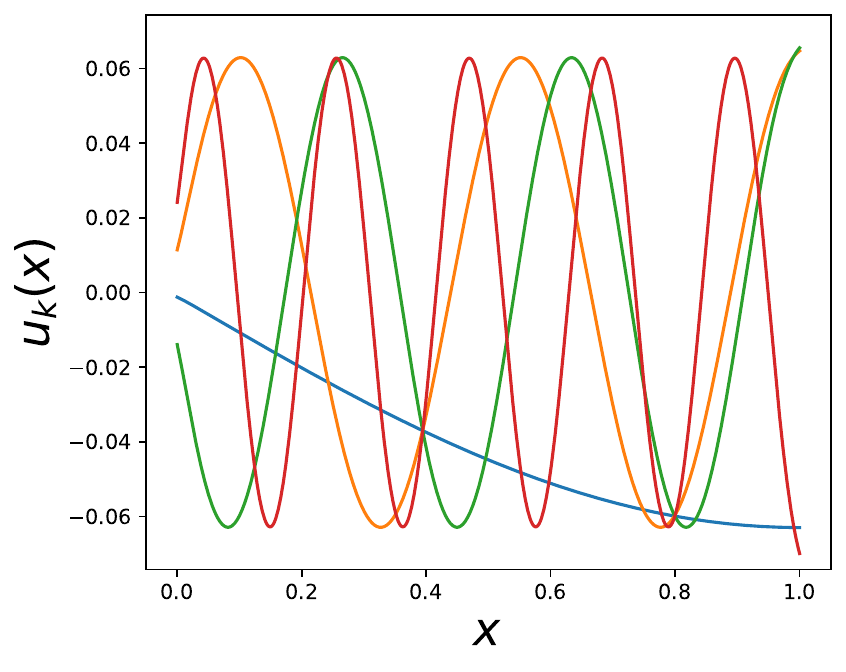}
        (b): tanh
    \end{minipage}%
    \begin{minipage}{.25\textwidth}
        \centering
        \includegraphics[width=\linewidth]{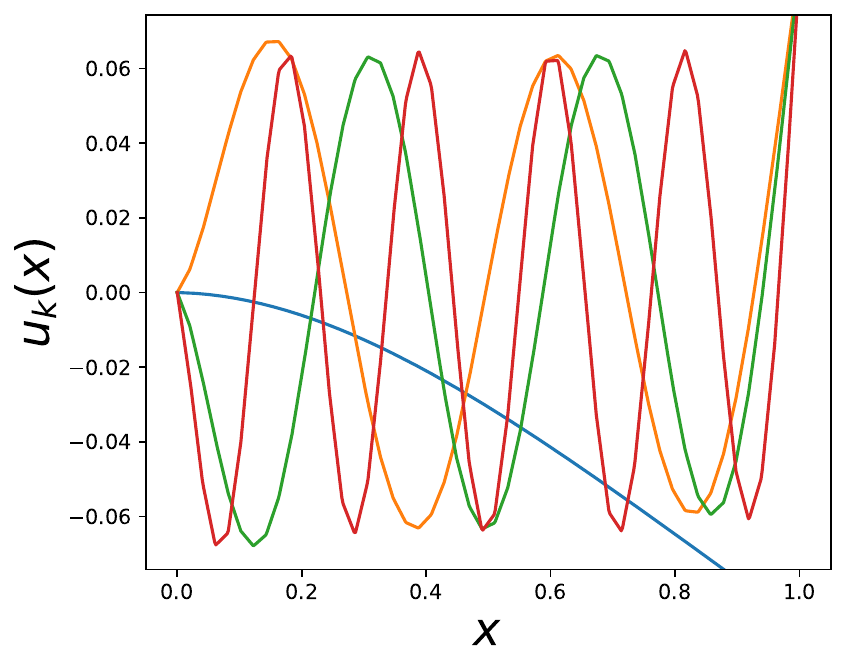}
        (c): ReLU
    \end{minipage}%
    \begin{minipage}{.25\textwidth}
        \centering
        \includegraphics[width=\linewidth]{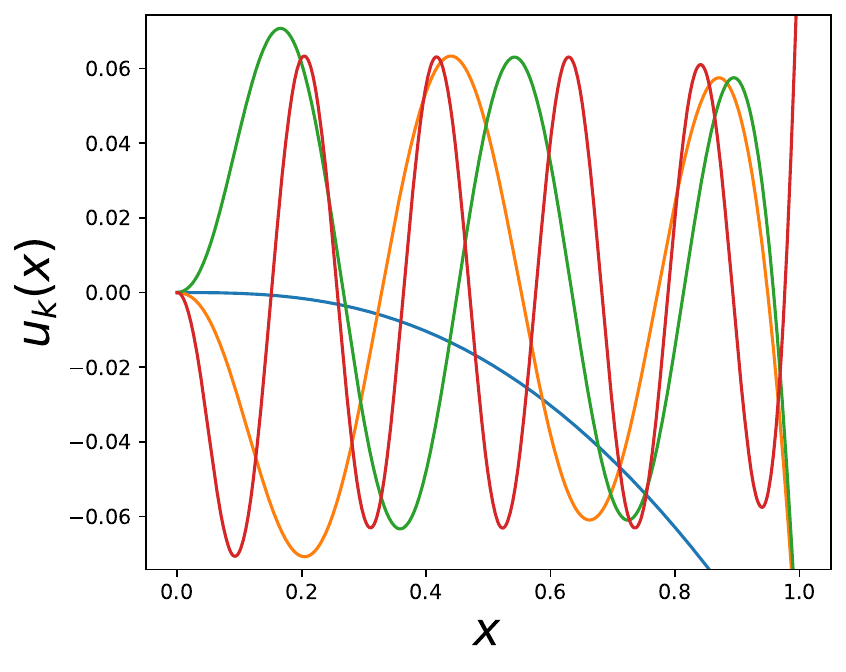}
        (d): ReLU2
    \end{minipage}%
    \caption{Visualization of principal components $u_{k}(x)$ of $\A$ for activation functions (a-d) where $k = [0,4,5,9]$ for $N = 5000$ and $M = 50$. In (a) we can see how unsuitable Heaviside is for practical problems where one must generalize smoothly across samples. In contrast, the tanh activation in (b) is able to preserve a smooth frequency basis while dealing with sparse sample observations. The (c) ReLU and (d) ReLU2 activations also exhibit smooth performance, but do not adhere as strictly to a frequency basis. Unlike tanh, however, ReLU and ReLU2 are not scale-sensitive so can always guarantee a smooth basis irrespective of the choice of $\sigma$.}
    \label{Fig:basis_50}
\end{figure*}

In Fig.~\ref{Fig:basis_50} we now want to inspect the principal components when we reduce the width $M$ of the network. Here we shall choose $M = 50$ and $N = 5000$. In (a) we can see an immediate problem with the Heaviside activation as it essentially employs a ``nearest neighbor'' style interpolated frequency basis. This is an undesirable property for smooth signal reconstruction due to the significant discontinuities at each observed sample. tanh fares substantially better in (b) forming a smoother DST frequency basis when using finite width. Again, the choice of scaling is important. For (a) ReLU, and (b) ReLU2 smooth bases are also obtained but again do not strictly adhere to a frequency basis. An obvious drawback to tanh, however, is that its smooth performance is scale sensitive. An incorrect choice of scaling will revert tanh back to the nearest neighbor interpolation observed for Heaviside. ReLU and ReLU2 are scale equivariant, enjoy smooth bases for finite width, but do not strictly adhere to a frequency basis. Visualizations of activation bases for GELU and SiLU exhibit similar properties to ReLU.

\begin{figure*}[!htb]
    \centering
    \begin{minipage}{.5\textwidth}
        \centering
        \includegraphics[width=\linewidth]{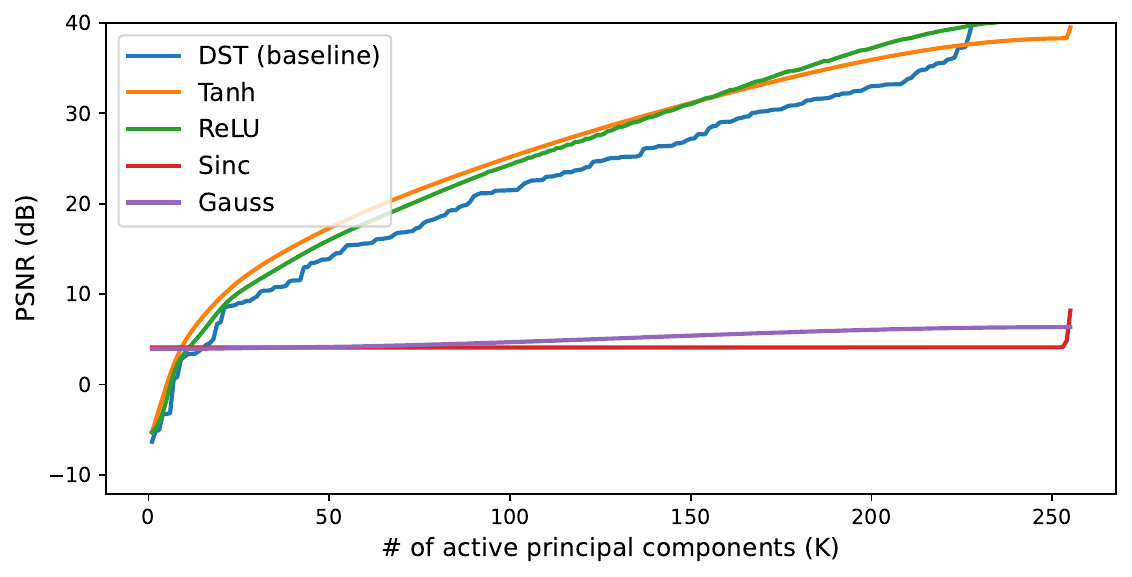}
        (a): $q = s_{\max} \cdot s^{-1}_{K}$, $\sigma = 1024$
    \end{minipage}%
    \begin{minipage}{.5\textwidth}
        \centering
        \includegraphics[width=\linewidth]{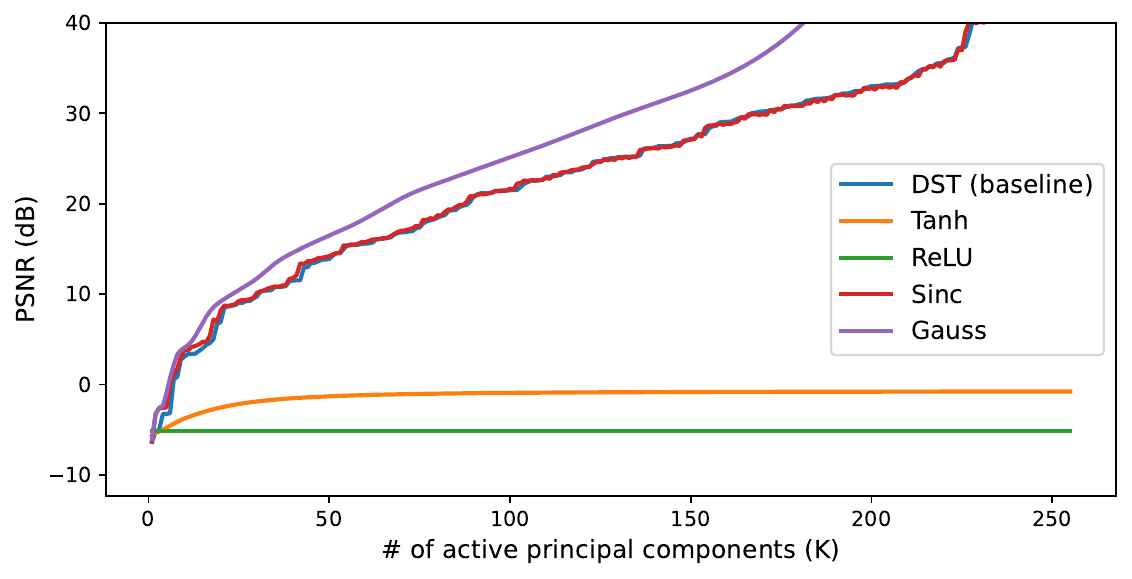}
        (b): $q = 100$, $\sigma = K$
    \end{minipage}%
    \caption{Reconstruction results from the peppers image. In (a) we are adaptively changing the number of iterations $q = s_{\max} \cdot s^{-1}_{K}$ in GD as a function of the desired number of principal components $K$ to preserve, while keep the scaling $\sigma = 1024$ fixed. We can observe that monotonic activations such as tanh and ReLU respond well (matching the DST baseline) to this type of regularization, whereas non-monotonic activations such as sinc and Gauss do not. In (b) we now keep the number of iterations $q = 100$ fixed, while adaptively changing the scaling $\sigma = K$. In this scenario performance is reversed with non-monotonic activations sinc and Gaussian matching the DST baseline, and monotonic activations tanh and ReLU having little to no regularization benefit.}
    \label{Fig:peppers}
\end{figure*}

\qsection{Signal reconstruction}
In Fig.~\ref{Fig:peppers} we attempt to reconstruct a row of the classic peppers image (row=$100$, of samples $N=256$, network width $M=1024$) using a shallow network that has its biases $\b$ equally spaced between zero and one. Our experiments are aimed at both monotonic (tanh, ReLU) and non-monotonic (sinc, Gaussian) activations. 
Analysis of Heaviside and ReLU2 activations is omitted due to their poor interpolation, and iteration efficiency respectively. We also include a discrete sine transform (DST) baseline $y(x) \approx \sum_{k=0}^{K-1} \gamma_{k} \cdot u_{k}(x)$ that uses the first $K$ sinusoid bases $u_{k}(x) = \sin(\pi \cdot [k + 0.5] \cdot x)$ to reconstruct the signal. The DST baseline does not use GD, but instead uses a closed form solution. On the $y-$axis we have the reconstruction performance measured in PSNR, and on the $x-$axis we have the number active principal components we are trying to preserve through regularization. For all our experiments the iteration efficient learning rate $\alpha = s^{-2}_{\max}$ is employed. 

In (a) we apply a GD regularization strategy where the number of iterations is a function of the desired principal components to keep active such that $q = s_{\max} \cdot s^{-1}_{K}$ (see Proposition~\ref{Prop:q}). We set the scaling factor of the activation to match the width of the network $\sigma = M = 1024$. Inspecting the results, one can see the monotonic activations closely match the performance of the DST baseline. This validates the role of varying $q$ as an implicit regularization mechanism for monotonic activations such as tanh \& ReLU. For non-monotonic, however, the results are quite poor with the varying of $q$ offering no regularization benefit whatsoever. Our previous analysis of the normalized singular value spectrum in Fig.~\ref{Fig:spectrum} for (sinc, Gaussian) predicted this as the singular value drop-off is tepid making it difficult for GD to effectively mask principal components.

In (b) we explore a strategy of varying the scaling factor as a function of the desired principal components to keep active such that $\sigma = K$ (see~\ref{Eq:shannon}). We keep the number of iterations used with GD fixed and modest $q = 100$. Here we see almost the opposite effect. The non-monotonic activations track the DST baseline well, with sinc offering almost identical performance. For monotonic, there is little to no benefit in adjusting the scaling factor. This result is obvious for ReLU since it is scale equivariant, but also makes sense for tanh as the activation will revert back to Heaviside as $\sigma \rightarrow \infty$. In this sense Heaviside acts as an upper-bound for tanh as the scale factor is varied. The spectral distribution in Fig.~\ref{Fig:spectrum} also explains why tanh enjoys a higher PSNR than ReLU, as the less rapid drop-off allows GD to keep active more principal components.    
\section{Discussion}
Based on the theoretical and empirical analysis put forward thus far, we can now make a key insight about the implicit spectral bias of GD within a shallow network. Inspecting Fig.~\ref{Fig:peppers} one can see that monotonic networks -- at least empirically -- exhibit a strikingly similar performance to a sinc network if suitably regularized. We make this connection more formal by realizing that the DST basis $u_{k}(x) = \sin(\pi \cdot [k + 0.5] x)$ can be re-written as $u(x;\omega) = \sin(2\pi \cdot \omega \cdot x)$ where $\omega$ refers to the frequency of the basis. The DST relationship is strongest for $r=1$ activations (Heaviside, tanh), but is approximately present for other activations $r > 1$ (ReLU, GELU, SiLU, etc.). We can then establish that the bandwidth $B$ (i.e. highest frequency component) of the signal is equal to $K = 2B - 0.5$. Due to the slow-transition of the GD masking function $K \approx 2B$ is an acceptable approximation. 

\begin{tcolorbox}[colback=yellow!20!white, colframe=black!50]
\textbf{Key Insights} 

1. an explicit relationship exists between GD's (learning rate $\alpha$, number of iterations $q$) and the number of active principal components $K$ -- see Proposition~\ref{Prop:q}. 

2. the principal components of monotonic activations adhere approximately to the discrete sine transform (DST) -- see Figs~\ref{Fig:basis_500} and~\ref{Fig:basis_50} -- which is bound by a maximum frequency $\omega_{\max} \leq B$. 

The following relationship can therefore be made $q \approx \alpha^{-2} \cdot s_{2B}$ -- for monotonic activations -- where $B$ is the desired bandwidth of the reconstruction.   
\end{tcolorbox}

\qsection{Bias \& Bandwidth}
Throughout this paper we have assumed we are only learning the weight vector $\w$, not the bias term $\b$. Shannon \& Whitaker's sampling theorem (see Eq.~\ref{Eq:shannon}) -- when using the sinc activation -- offers guidance on how to distribute~$\b$ for a desired bandwidth. The theorem tells us that equal $\sigma^{-1}$ spaced biases are optimal. It also tells us that if we entertain spacings that are smaller, we do not gain anything in terms of preserving more of the signal. This insight is important, as it demonstrates that shallow sinc networks -- unlike monotonic -- are unable to memorize arbitrary signals if the spacings between samples is below $\sigma^{-1}$ irrespective of the width or distribution of $\b$. It also provides a useful heuristic for the minimum width $M \geq \sigma$ of the network if one knows the desired bandwidth of the reconstruction in advance. 

\qsection{Multiple Dimensions}
Our discussion thus far has focused only on single dimension coordinates $x \in \mathbb{R}$. We can also explore the generalization to multiple dimensions $\x \in \mathbb{R}^{D}$. In Appendix~\ref{App:MD} we offer up visualization of the principal components for the case of $D = 2$. Unlike for the 1D case, the index of the principal component no longer have any relationship to bandwidth or a spectral basis. In general, however, there is a strong relationship between the ordering of the singular values and the smoothness of the principal component (which is consistent with the 1D analysis offered in this paper).

\qsection{Deepness \& Future Work}
The focus of this work was to establish a better understanding of the implicit spectral bias offered by GD within a 1D shallow network. 
The insights gained from the 1D shallow case have clear ramifications to real-world applications that involve signals $y(\x)$ of much higher dimensions and also the use of deep rather than shallow networks. The insight that GD can be viewed as a shrinking operator on the singular values of a network's Jacobian could open up new avenues of investigation for regularizing deep networks. The realization that there exists a class of non-monotonic activations that are regularized solely through scaling (not through iterations or learning rate) gives new hope for designing architectures that are super-efficient to train but can still generalize. Finally, the establishment of a connection between spline smoothing/regression and monotonic activations may open up new strategies for better explaining the foundations of generalization properties of modern deep networks. 

{
    \small
    \bibliographystyle{ieeenat_fullname}
    \bibliography{refs}
}

\clearpage
\maketitlesupplementary
\appendix

\section{Activation as a spline regularizer/smoother}
\label{App:hside}
Here we want to empirically validate a connection between activation choice in a shallow network and spline regularization/smoothing. Let us rewrite Eq.~\ref{Eq:spline_error} discretely, 
\begin{equation}
    \arg \min_{\f} \frac{1}{2}||\y - \mathbf{D}\f]||_{2}^{2} + \frac{\lambda}{2} ||\nabla^{r} \f||_{2}^{2}
    \label{Eq:num_spline_error}
\end{equation}
$\f$ is a $M$ dimensional discrete approximation to the continuous function $f$ and $\y = [y(\hat{x})]_{\hat{x} \in \mathbb{D}}$ are the $N$ discrete signal samples. The matrix $\mathbf{D}$ is a $N \times M$ concatenated matrix of indicator vectors corresponding to the training coordinates $\hat{x} \in \mathbb{D}$. We approximate the continuous function $\frac{d}{dx} f(x)$ as $\nabla \f$ where $\nabla$ is a $M \times M$ finite difference convolutional matrix using filter weights $[\Delta x,- \Delta x]$ where $\Delta x = M^{-1}$ and $x \in [0,1]$. The value $r$ denotes the order of the gradients being minimized. As $M \rightarrow \infty$ the numerical approximation of the continuous integral in Eq.~\ref{Eq:spline_error} becomes increasingly accurate. 

\qsection{Shifting through convolution}
Here we can entertain different types of discrete convolution when forming $\nabla$, specifically \emph{valid}, \emph{same} and \emph{circulant} all making different assumptions about the boundary conditions of the signal. Valid convolution in many ways is the most instructive as it is dealing only with the finite differences of the signal itself, ignoring the vagaries of signal boundary assumptions. However, it does not provide square $N \times N$ convolutional matrices. We instead employed same convolution in our experiments due to its simple handling of signal boundary conditions and its ability to produce a square non-singular $\nabla$.

\begin{figure*}[!htb]
    \centering
    \begin{minipage}{.33\textwidth}
        \centering
        \includegraphics[width=\linewidth]{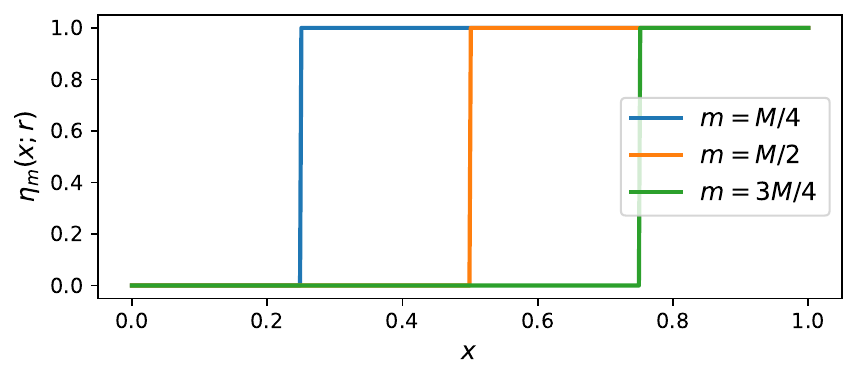}
        (a): $r=1$
    \end{minipage}%
    \begin{minipage}{.33\textwidth}
        \centering
        \includegraphics[width=\linewidth]{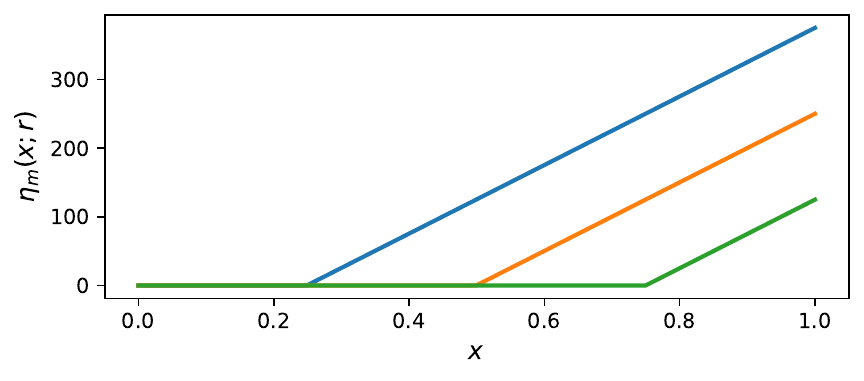}
        (b): $r=2$
    \end{minipage}%
    \begin{minipage}{.33\textwidth}
        \centering
        \includegraphics[width=\linewidth]{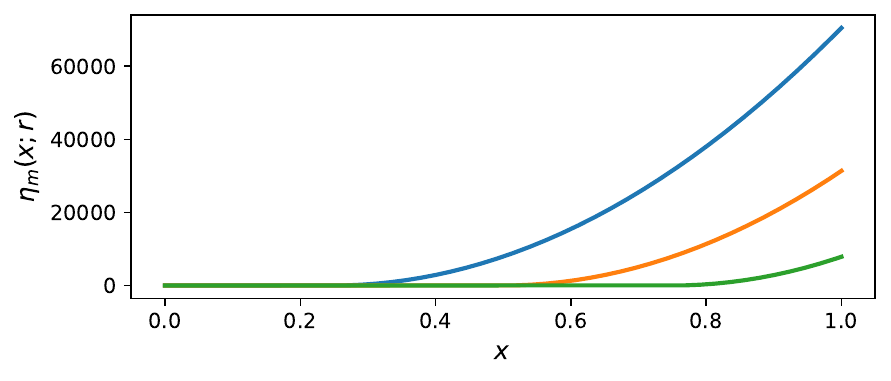}
        (c): $r=3$
    \end{minipage}%
    \caption{Visualization of column vectors of $\nabla^{-r} = [\boldsymbol{\eta}^{r}_{0},\ldots,\boldsymbol{\eta}^{r}_{M-1}]$. We approximate a continuous function by setting $\eta_{m}(x;r) = \boldsymbol{\eta}^{r}[x \cdot M \cdot T^{-1}]$. For (a) one can see the column vectors strike a strong resemblance to the shifted Heaviside function $\eta_{m}(x;1) = [(x - b_{m}) > 0]$. In (b) the function matches the shifted ReLU function $\eta_{m}(x;2) = x \cdot [(x - b_{m}) > 0] = \max(x - b_{m},0)$. Finally, in (c) we see the shifted ReLU2 function $\eta_{m}(x;3) = x^{2} \cdot [(x - b_{m}) > 0] = \max(x - b_{m},0)^{2}$. The bias term is defined as $b_{m} = m \cdot M^{-1} \cdot T$. For our visualization we set $M = 5000$, and visualize the column vectors at $m = [M/4, M/2, 3M/4]$.} 
    \label{Fig:nabla}
\end{figure*}

\qsection{Visualizing $\nabla^{-r}$}
Assuming $\nabla^{r}$ is non-singular and square we can rewrite Eq.~\ref{Eq:num_spline_error} using $\f = \nabla^{-r} \w$ 
\begin{equation}
    \arg \min_{\w} \frac{1}{2}||\y - \mathbf{D} \nabla^{-r} \w]||_{2}^{2} + \frac{\lambda}{2} ||\w||_{2}^{2} \;\;. 
    \label{Eq:num_shallow_error}
\end{equation}

We can visualize the row vectors of the $\nabla^{-r}$ for: (a) $r=1$, (b) $r=2$, and (c) $r=3$ in Fig.~\ref{Fig:nabla}. We depict $\nabla^{-r} = [\boldsymbol{\eta}^{r}_{0},\ldots,\boldsymbol{\eta}^{r}_{M-1}]^{T}$ as a concatenation of column vectors. In our visualizations we shall approximate the discrete vectors as continuous functions $\eta_{k}(x;r) \approx \boldsymbol{\eta}^{r}_{k}[x \cdot M \cdot T^{-1}]$. 

For (a) we found the visualization closely resembles the shifted Heaviside function $\eta_{m}(x;1) = [(x - b_{m}) > 0]$. In (b) it matches the shifted ReLU function $\eta_{m}(x;2) = x \cdot [(x - b_{m}) > 0] = \max(x - b_{m},0)$. Finally, in (c) it matches the shifted ReLU2 function $\eta_{m}(x;3) = x^{2} \cdot [(x - b_{m}) > 0] = \max(x - b_{m},0)^{2}$. The bias shift is defined as $b_{m} = m \cdot M^{-1} \cdot T$. We visualize for shifts $m = [M/4, M/2, 3M/4]$ where $M = 5000$ and $T =1$. This result empirically validates our claims in Proposition~\ref{Prop:Act} that $\eta(x;r) = x^{r-1} \cdot [x > 0]$. With same convolution $\nabla$ is not a symmetrical matrix therefore only positive discrete values of $r \in \mathbb{Z}^{+} \geq 1$ can be entertained.      

\section{GD as a shrinkage operator}
\label{App:GD}
Gradient descent (GD) can be used to great effect to implicitly regularize least-squares problems. Applying the iterative nature of GD described in Eq.~\ref{Eq:GD1} we obtain, 

\begin{equation}
\w_{k+1} \rightarrow \w_{k}
 - \alpha \cdot \frac{\partial E(\w)}{\partial \w^{T}}
\end{equation}
where $k$ is the current iteration.

Using the definition for $\frac{\partial E(\w)}{\partial \w^{T}}$ (see Eq.~\ref{Eq:GD2}) and assuming $\mathbf{w}_{0} = \mathbf{0}$ we then have,

\begin{equation}
\mathbf{w}_{1} = \alpha \mathbf{A}^{T} \mathbf{y}
\end{equation}

\begin{equation}
\mathbf{w}_{2} = (\mathbf{I} - \alpha \mathbf{A}^{T} \mathbf{A}) \alpha \mathbf{A}^{T} \mathbf{y} + \alpha \mathbf{A}^{T} \mathbf{y}
\end{equation}

\begin{equation}
\mathbf{w}_{3} = (\mathbf{I} - \alpha \mathbf{A}^{T} \mathbf{A})^{2} \alpha \mathbf{A}^{T} \mathbf{y} + (\I - \alpha \mathbf{A}^{T} \mathbf{A}) \alpha \mathbf{A}^{T} \mathbf{y} + \alpha \mathbf{A}^{T} \mathbf{y}.
\end{equation}

This can be generalized using the Neumann series,
\begin{equation}
\sum_{k=0}^{q-1} z^{k} = (1 - z^{q})(1 - z)^{-1}
\end{equation}

such that,

\begin{eqnarray}
\mathbf{w}_{q} & = & \sum_{k=0}^{q-1}(\mathbf{I} - \alpha \mathbf{A}^{T} \mathbf{A})^{k-1} \alpha \mathbf{A}^{T} \mathbf{y} \\
& = & [\mathbf{I} - (\mathbf{I} - \alpha \mathbf{A}^{T} \mathbf{A})^{q}](\alpha \mathbf{A}^{T} \mathbf{A})^{-1} \alpha \mathbf{A}^{T} \mathbf{y}
\end{eqnarray}
where $q$ is the total number of iterations. We can implement this as a shrinkage operator on the singular values of $\A = \U [\diag(\s)] \V^{T}$ such that,
\begin{equation}
\hat{s}^{-1} = [1 - (1 - \alpha \cdot s^{2})^{q}] \cdot s^{-1}
\label{Eq:GD_naive}
\end{equation}
thus resulting in the masking function in Eq.~\ref{Eq:mask_gd} where $\hat{s}^{-1} = m_{\tiny{\mbox{gd}}}(s;\alpha,q) \cdot s^{-1}$ and $m_{\tiny{\mbox{gd}}}(s;\alpha,q) = [1 - (1 - \alpha \cdot s^{2})^{q}]$. The framework easily extends to the more practical case where $\w_{0} \neq \mathbf{0}$ where we update $\y \rightarrow \y - \A \w_{0}$. This derivation is especially useful as it makes clear that $\alpha$ and $q$ are two distinct and important variables. In particular $\alpha \leq s^{-2}_{\max}$ to ensure non-catastrophic performance. A drawback to this particular derivation, however, is that it is susceptible to the vagaries of floating point precision on singular values that are not present in practical gradient descent. 

\qsection{Gradient flows}
Neural tangent kernel (NTK)~\cite{jacot2018neural} -- using gradient flows (GF) -- is a popular choice for approximating the effect of GD on a network $f(x)$ (assuming least-squares) such that, 
\begin{equation}
    \y_{q} = [\I - \exp(-\alpha \cdot \mathbf{K} \cdot q )] \y
    \label{Eq:NTK}
\end{equation}
and, 
\begin{equation}
    \mathbf{K} = [\nabla f(\x)]^{T} \nabla f(\x)
\end{equation}
such that $\y_{q}$ is the approximation to the signal $\y$ at iteration $q$ and learning rate $\alpha$. The analysis in our paper has focused on shallow networks (see Eq.~\ref{Eq:shallow}) so we can show that $f(\x) = \A \w$ and $\nabla f(\x) = \A^{T}$. We can then rewrite Eq.~\ref{Eq:NTK} as, 
\begin{equation}
    \y_{q} = \U [\I - \mbox{diag}(\exp[-\alpha \cdot \s^{2} \cdot q])] \U^{T}
\end{equation}
where $\A = \U [\mbox{diag}(\s)] \V^{T}$ is the SVD decomposition. If we assume $\y_{q} = \A \w_{q}$ -- where $\w_{q}$ is the approximation to $\w^{*}$ at iteration $q$ -- we can now obtain, 
\begin{equation}
    \w_{q} = \V \mbox{diag}(\s)^{-1} [\I - \mbox{diag}(\exp[-\alpha \cdot \s^{2} \cdot q])]  \U^{T} \;\;. 
\end{equation}
We can also implement this as a shrinkage operator on the singular values of $\A$ such that
\begin{equation}
\hat{s}^{-1} = [1 - \exp(-t \cdot s^{2})] \cdot s^{-1}
\end{equation}
thus validating Proposition~\ref{Prop:GD} where $\hat{s}^{-1} = m_{\tiny{\mbox{gd}}}(s;t) \cdot s^{-1}$, $m_{\tiny{\mbox{gd}}}(s;t) = [1 - \exp(-t \cdot s^{2})]$and $t = \alpha \cdot q$. Unlike Eq.~\ref{Eq:GD_naive} the gradient flows derivation is not sensitive to floating point precision, resulting in a near perfect approximation to GD. In practice, however, the learning rate is still bounded by the maximum singular value such that $\alpha \leq s^{-2}_{\max}$ as demonstrated in our naive derivation.

\section{Measuring active window width $\rho$}
\label{App:rho}
Using Eq.~\ref{Eq:mask_gd} we can estimate the singular value in the masking function where the value drops to $m(s;\alpha,q) = 1 - \epsilon$ such that, 
\begin{equation}
s(\epsilon) = \sqrt{\frac{-\log(\epsilon)}{\alpha \cdot q}} \;\;. 
\end{equation}
We can then define the active window as $\rho(\epsilon) = s(\epsilon) \cdot s^{-1}_{\max}$ which simplifies down to,  
\begin{equation}
\rho(\epsilon) = \sqrt{\frac{q}{-\log(\epsilon)}} \;\;. 
\end{equation}
where we set the learning rate $\alpha = s^{-2}_{\max}$ to ensure optimal iteration efficiency. Here we can see that $\rho \propto \sqrt{q}$, and becomes equal ($\rho = \sqrt{q}$) when $\epsilon = \exp(-1)$ thus valdating Proposition~\ref{Prop:q}. 

\section{Iteration Efficiency of GD}
\label{App:iter}
\begin{figure}[!htb]
    \centering
    \includegraphics[width=\linewidth]{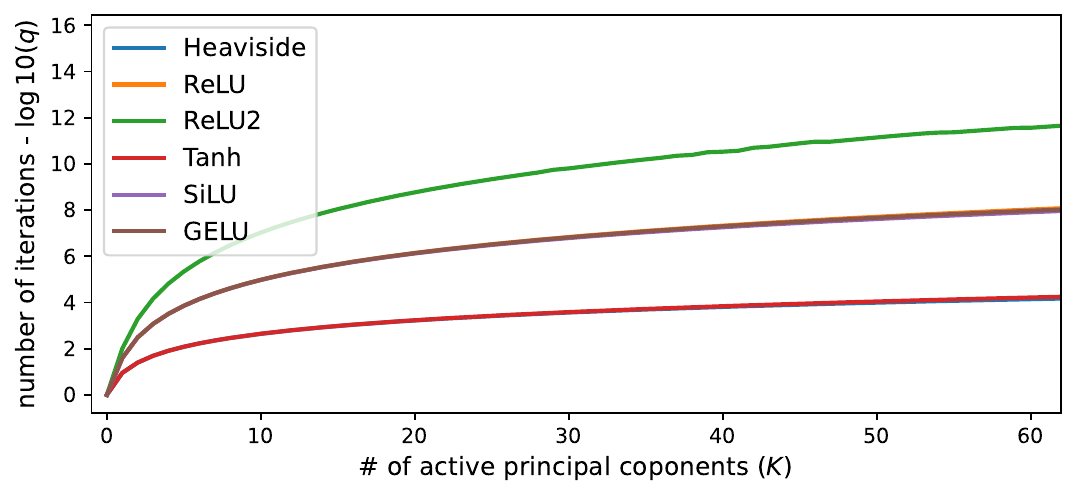}
    \caption{Visualization of the number of iterations $q$ required to preserve $K$ active principal components using the relation from Proposition~\ref{Prop:q}. $r=1$ activations (Heaviside, tanh) are orders of magnitude more efficient compared to modern activations (ReLU, GELU, SiLU, ReLU2).}
    \label{Fig:iter}
\end{figure}

Using the relation from Proposition~\ref{Prop:q} (see Eq.~\ref{Eq:q}) we can visualize the iteration efficiency of GD with respect to the number of active principal components. In Fig.~\ref{Fig:iter} one can see the number of iterations required $q$ on the $y-$axis, and the number of active principal components $K$ on the $x-$axis. $r=1$ activations (Heaviside, tanh) exhibit a significant efficiency advantage over more conventional activations. $r=2$ activations (ReLU, GELU, SiLU) require (in some instances) almost three orders of magnitude more iterations. $r=3$ activations (ReLU2) fare even worse, requiring almost six orders of magnitude more iterations.  

\section{Multi-Dimensional Signals}
\label{App:MD}
\begin{figure}[!htb]
    \centering
    \includegraphics[width=\linewidth]{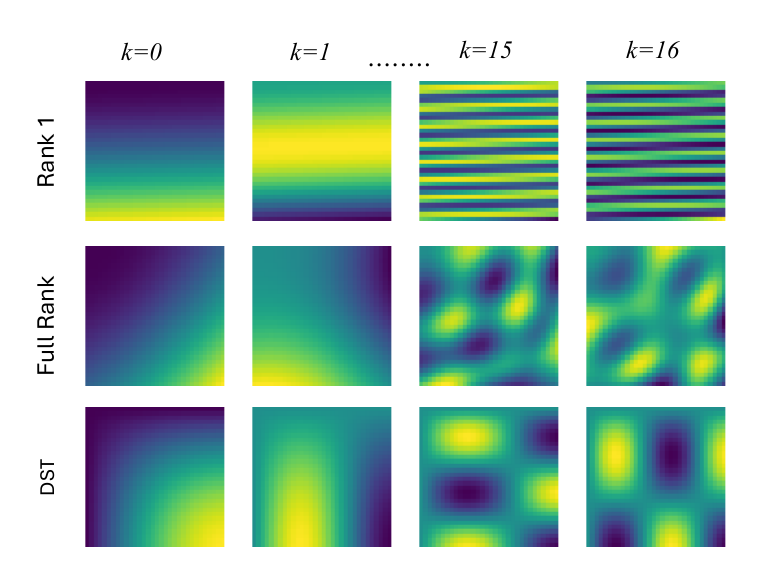}
    \caption{We visualize 2D principal components $u_{k}(x,y)$ from $\mathbf{A}$ (using a ReLU activation) for indexes $k = [0,1,15,16]$. For row 1 we see the principal components for Zhang et al.'s rank 1 initialization of $\V$ (see Eq.~\ref{Eq:ND}). One can see for this initialization of $\V$ this principal components violate the 2D smoothness assumption exhibiting variation only in the horizontal direction. Row 2 depicts the full rank initialization of $\V$ --  exhibiting improved bidirectional smoothness. Row 3 shows the 2D DST basis.}
    \label{fig:DST}
\end{figure}

Our discussion thus far has focused only on single dimension coordinates $x \in \mathbb{R}$. Here we want to explore the generalization to multiple dimensions $\x \in \mathbb{R}^{D}$. We can write this in the general form, 
\begin{equation}
    f(\x) = \w^{T} \eta(\V \x - \b)
    \label{Eq:ND}
\end{equation}
such that $\V = [\v_{0},\ldots,\v_{M-1}]^{T}$ is a concatenation of vectors $\v \in \mathbb{R}^{D}$. For the multidimensional case we can reframe the linear model $\y = \A \w$ (see Eq.~\ref{Eq:Ew}) as $\y = [y(\hat{\x})]_{\hat{\x} \in \mathbb{D}}$, $\A = [\eta(\v_{j}^{T} \hat{\x}_{i} - b_{j})]_{ij}$ as an $N \times M$ matrix where $N = |\mathbb{D}|$. Zhang et al.~\cite{zhang2017understanding} trivially set $\V = \mathbf{1} \v^{T}$ -- for the multidimensional case of ReLU memorization -- to be a rank one matrix as a $\v$ can always be chosen so that $\v^{T} \x$ is unique (so long as $\x$ itself is unique)\footnote{For a 2D discrete signal with $r$ rows and $c$ columns we can use $\v = [r,1]^{T}$ or $\v = [1,c]^{T}$ to ensure a unique mapping of the 2D coordinates into 1D. In our experiments we used $\v = [r,1]^{T}$ (see row 1 in Fig.~\ref{fig:DST}).}.  

In row 1 of Fig.~\ref{fig:DST} we can visualize the 2D principal components $\u_{k}(x,y)$ of $\A$ using Zhang et al.'s rank one initialization strategy. Unlike for the 1D case, the index of the principal component no longer has any relationship to bandwidth or the DST basis. In fact one can see that there is no longer any relationship between the ordering of the singular values and the smoothness of the principal component. This demonstrates that although Zhang et al.'s strategy is a useful for analytically proving ReLU memorization it has a catastrophic effect on generalization -- which is one of the central practical reasons for using the ReLU activation. Things get noticeably better in row 2 of Fig.~\ref{fig:DST} when we use a full rank initialization of $\V$ as the relation between singular value order and smoothness returns -- further rationalizing the common practice of using a full rank initialization of $\V$. 

\end{document}